\documentclass{article}

\PassOptionsToPackage{numbers, compress}{natbib} %

\usepackage[preprint]{neurips_2026}

\usepackage[utf8]{inputenc} %
\usepackage[T1]{fontenc}    %
\usepackage{hyperref}       %
\usepackage{url}            %
\usepackage{booktabs}       %
\usepackage{amsmath}        %
\usepackage{amsfonts}       %
\usepackage{nicefrac}       %
\usepackage{microtype}      %
\usepackage{xcolor}         %
\usepackage{xspace}         %
\usepackage{tikz}           %
\usepackage{algorithm}      %
\usepackage[noEnd=true,indLines=true,italicComments=true]{algpseudocodex} %

\usepackage{fvextra}
\usepackage[breakable, skins]{tcolorbox}

\usepackage{xcolor}
\usepackage{tikz}
\usetikzlibrary{positioning,arrows.meta,fit,calc,matrix,backgrounds}
\newcommand{\alg}{\textsc{Cantante}\xspace}

\DeclareUnicodeCharacter{2212}{-}
\newcounter{prompt}
\renewcommand{\theprompt}{\arabic{prompt}}

\newcommand{\promptboxfile}[3][]{%
    \refstepcounter{prompt}%
    \ifx&#1&\else\label{#1}\fi%
    \begin{tcolorbox}[
        title={Prompt \theprompt: #2},
        colback=blue!4!white,
        colframe=gray!40,
        colbacktitle=gray!20!black,
        coltitle=white,
        fonttitle=\bfseries\small,
        breakable,
        enhanced,
        boxrule=0.5pt,
        arc=2pt,
        left=6pt, right=6pt, top=4pt, bottom=4pt,
        drop shadow,
    ]
    \VerbatimInput[breaklines=true, breakanywhere=true, fontsize=\small, baselinestretch=1.1]{#3}
    \end{tcolorbox}%
}

\newcommand{\Nagents}{N}                         %
\newcommand{\Nconfigs}{K}                        %
\newcommand{\Niters}{T}                          %

\newcommand{\agentidx}{a}                        %
\newcommand{\configidx}{i}                       %
\newcommand{\queryidx}{q}                        %

\newcommand{\graph}{\mathcal{G}}                 %
\newcommand{\nodes}{\mathcal{V}}                 %
\newcommand{\edges}{\mathcal{E}}                 %
\newcommand{\pred}{Y}                          %

\newcommand{\queryset}{Q}                         %
\newcommand{\querydist}{\mathcal{D}}             %

\newcommand{\agent}[1]{A_{#1}}                   %
\newcommand{\agentparams}[1]{\theta_{#1}}        %
\newcommand{\agentin}[1]{x_{#1}}                 %
\newcommand{\agentout}[1]{y_{#1}}                %
\newcommand{\prompt}[1]{p_{#1}}                  %
\newcommand{\agentcredit}[1]{c_{#1}}             %

\newcommand{\param}[2]{\theta_{#1,#2}}           %
\newcommand{\paramspace}[1]{\Theta_{#1}}         %
\newcommand{\optimizer}[1]{O_{#1}}               %
\newcommand{\jointconfig}[1]{\Phi_{#1}}             %
\newcommand{\group}{G}                   %
\newcommand{\groupsize}{g}         %
\newcommand{\trajectory}[2]{\xi_{#1,#2}}        %
\newcommand{\trajectoryplane}{\xi}        %
\newcommand{\score}[2]{s_{#1,#2}}                %
\newcommand{\sysscore}{S}                        %
\newcommand{\credit}[3]{c_{#1,#2,#3}}            %
\newcommand{\aggcredit}[2]{r_{#1,#2}}            %
\newcommand{\task}{\tau}                         %

\newcommand{\Suggest}{\textsc{suggest}}
\newcommand{\Update}{\textsc{update}}
\newcommand{\Evaluate}{\textsc{evaluate}}
\newcommand{\Attribute}{\textsc{attribute}}

\newcommand{\attributer}{\textsc{Attributer}}

\definecolor{cbOrange}{HTML}{D55E00}
\definecolor{cbGreen}{HTML}{009E73}

\DeclareMathOperator*{\argmax}{arg\,max}

\newcommand{\githuburl}{\url{https://github.com/finitearth/cantante}}

\tikzset{
  agent/.style={
    draw=violet!80,
    fill=violet!12,
    rounded corners=3pt,
    minimum height=7mm,
    minimum width=22mm,
    inner sep=3pt,
    align=center,
    font=\sffamily\scriptsize
  },
  io/.style={
    draw=violet!80,
    fill=violet!28,
    rounded corners=9pt,
    minimum height=7mm,
    minimum width=15mm,
    inner sep=3pt,
    align=center,
    font=\sffamily\scriptsize
  },
  flow/.style={-{Stealth[length=1.8mm,width=1.2mm]}, thick, draw=black!75},
  cond/.style={-{Stealth[length=1.8mm,width=1.2mm]}, thick, dashed, draw=black!60},
  tool/.style={
    draw=violet!80, fill=violet!28,
    circle, outer sep=0pt, inner sep=0pt,
    minimum size=3.5mm, font=\sffamily\tiny\bfseries
  }
}

\usepackage{graphicx}
\usepackage{caption}
\usepackage{subcaption}
\usepackage{wrapfig}
\usepackage{paralist}
\usepackage{siunitx}

\DeclareSIUnit{\pp}{pp}

\title{\alg: Optimizing Agentic Systems via\\Contrastive Credit Attribution}

\author{%
  Tom Zehle \\
  University of Freiburg\\
  ELLIS Institute Tübingen\\
  \texttt{tom.zehle@tue.ellis.eu} \\
}

\begin{document}

\maketitle

\begin{abstract}
LLM-based multi-agent systems have demonstrated strong performance across complex real-world tasks, such as software engineering, predictive modeling, and retrieval-augmented generation. Yet automating their configuration remains a structural challenge, as scores are available only at the system level, whereas the parameters governing agent behavior are local. We argue that optimizing these systems is fundamentally a credit-assignment problem.
We therefore introduce \alg, a framework that decomposes system-level rewards into per-agent update signals by contrasting rollouts of multiple joint configurations on the same query. We instantiate it for prompt optimization, treating agent prompts as learnable system parameters.
We evaluate \alg against GEPA and MIPROv2 on programming (MBPP), mathematical reasoning (GSM8K), and multi-hop question answering (HotpotQA). Across these benchmarks, \alg achieves the best average rank among all evaluated optimizers and consistently outperforms unoptimized prompts. It improves over the strongest baseline by $+18.9$ percentage points on MBPP and $+12.5$ percentage points on GSM8K, while incurring a lower inference cost. It remains within one standard deviation of the strongest baseline on HotpotQA. Crucially, our credit correlation analysis confirms that the attributer produces meaningful per-agent signals rather than echoing the global system score.
\end{abstract}

\section{Introduction}
\label{sec:intro}
\begin{wrapfigure}{r}{0.43\textwidth}
    \vspace*{-2em}
    \centering
    \includegraphics[width=0.42\textwidth]{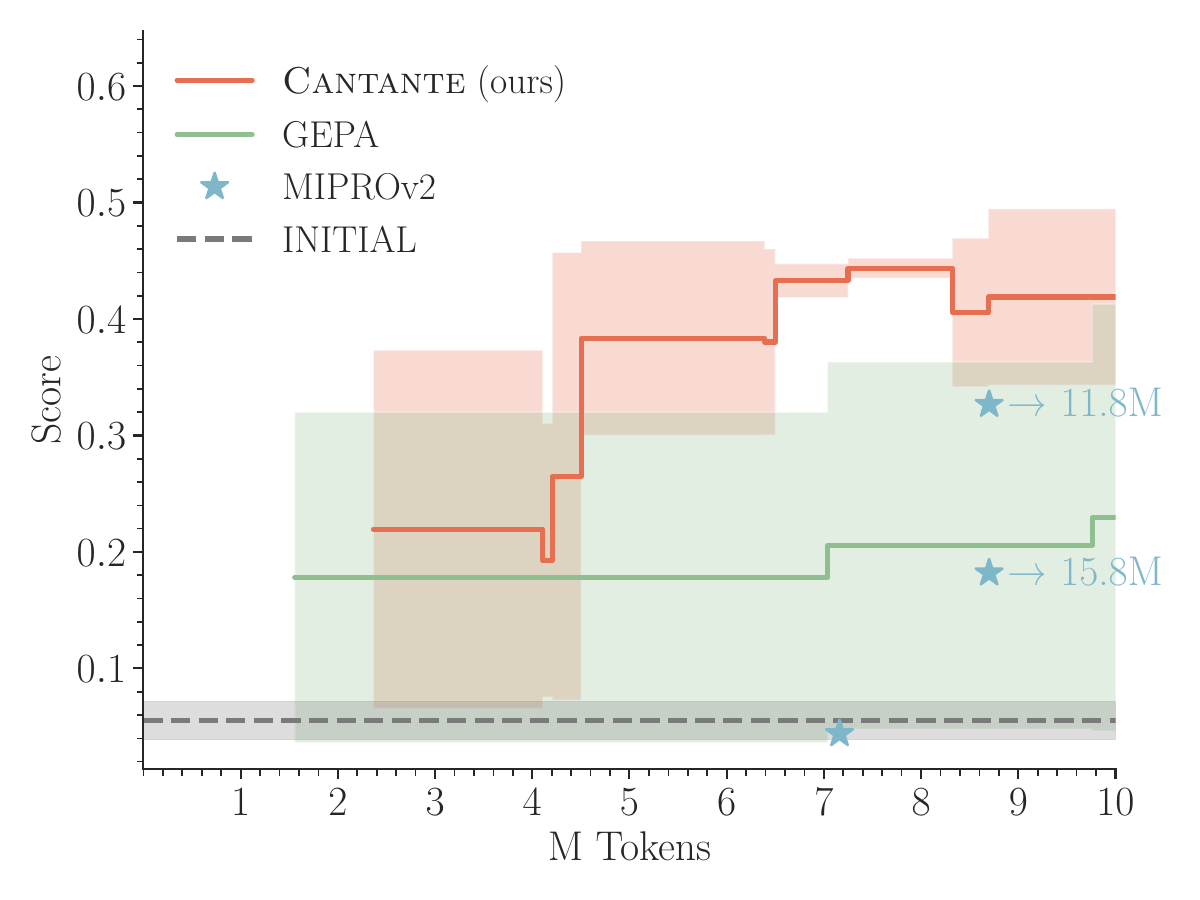}
    \caption{Trajectories on MBPP.\\\alg improves steadily, reaching the highest final accuracy.}%
    \label{fig:trajectory_mbpp}
    \vspace*{-1em}
\end{wrapfigure}

Agentic systems built on large language models have demonstrated strong empirical performance across a range of complex, real-world tasks, from autonomous software engineering \citep{yang2024swe} to end-to-end machine learning pipelines \citep{fang2026mlzero} and multi-hop retrieval \citep{chang-etal-2025-main}.
More broadly, multi-agent systems (MAS) turn LLMs from passive conversational models into autonomous problem-solving systems.
Yet realizing this potential in practice remains labor-intensive: each deployment requires manual selection of agents, hand-authored prompts, workflow design, and tool setup.

Manually tuning MAS configurations, however, contradicts the main paradigm of machine learning: Just as no practitioner hand-tunes neural network weights, manual iteration over agent prompts and roles is a problem that should be solved via optimization.
The manual labor required grows combinatorially with each agent requiring its own configuration and each prompt requiring a co-adaption to the behavior of the others, making trial-and-error tuning both tedious and ineffective.
Automating the configuration of MAS would reduce this burden and enable principled adaptation to new tasks and models, but doing so is non-trivial.
Not only is the search space high-dimensional, but more fundamentally, the optimization problem has a structural mismatch: agent-level parameters are \textit{local}, while evaluation feedback is \textit{global}, available only as a scalar signal over the full workflow output.

We argue that \textit{multi-agent system optimization is fundamentally a credit-assignment problem}: global feedback must be converted into per-agent update signals. In a planner--coder--evaluator workflow, for example, faulty code introduced by the coder should not trigger equally negative updates in the planner and evaluator. This problem is well-established in multi-agent reinforcement learning, where cooperative agents must each receive a learning signal derived from a shared global reward~\citep{wolpert2001optimal, coma}. Existing approaches to LLM-based MAS optimization largely sidestep this issue by propagating global scores as update signals to individual agents~\citep{agrawal2025gepareflectivepromptevolution, zhou2025multi}.

We therefore introduce \alg{}, a novel framework for node-level optimization in MAS based on system-level rewards and reasoning trajectories.
The central mechanism is \textit{contrastive in-group attribution}: by comparing rollouts of multiple joint configurations for the same query, the attribution model isolates each agent's parameterization's contribution to the observed differences in outcomes. This method is independent of the choice of local optimizer.
We instantiate this framework for prompt optimization, treating agent prompts as learnable system parameters rather than fixed implementation details.

We evaluate \alg{} on workflow graphs with non-trivial structure, including conditional edges, ensemble nodes, and tool-using agents, across programming (MBPP), mathematical reasoning (GSM8K), and multi-hop question answering (HotpotQA). \alg obtains the best average rank among all evaluated optimizers, with the largest gains occurring on MBPP and GSM8K, where \alg improves over the strongest baseline by 18.9 percentage points (pp) and \SI[round-precision=1]{12.5}{\pp}, respectively. The MBPP trajectory plot in Figure~\ref{fig:trajectory_mbpp} further shows that \alg reaches these gains steadily over optimization, rather than through a single late-stage jump. We show that these gains are achieved without substantially increasing inference-time token cost relative to the unoptimized prompts. On HotpotQA, \alg remains within a standard deviation of the best baseline and improves over the unoptimized initial configuration. Finally, our credit correlation analysis suggests that the attributer produces meaningful per-agent signals rather than merely echoing the global system score and that these signals can serve as a diagnostic tool for MAS topology design.

\paragraph{Contributions.} We make the following contributions:
\begin{inparaenum}[I)]
    \item We formalize the optimization of LLM-based multi-agent systems as a credit-assignment problem and introduce \alg{}. %
    \item We instantiate \alg{} for prompt optimization, treating agent prompts as learnable system parameters, and evaluate it against MAS-capable prompt optimizers across multiple benchmarks.
    \item We isolate contrastive attribution itself — rather than the underlying local optimizer — as the principal driver of \alg's gains via an identity-attribution ablation, and further demonstrate robustness to attributer prompt, attributer model, local optimizer choice, and dataset size.
    \item We release the full implementation of \alg{} to support adoption by practitioners and to enable further research on attribution-driven MAS optimization, available at \githuburl{}.
\end{inparaenum}

\section{Related Work}
\label{sec:related_works}
\paragraph{LLM-Based Multi-Agent Systems.}
Multi-agent systems consist of multiple autonomous agents that coordinate their actions to solve a shared task~\citep{wooldridge1995intelligent}.
LLM-based variants extend this paradigm through a language interface, composing agents via role specialization, communication, planning, memory, and tool use~\citep{guo2024large}.
Representative applications include autonomous software engineering with SWEAgent~\citep{yang2024swe}, automated predictive modeling with MLZero~\citep{fang2026mlzero}, and agent-based retrieval with MAIN-RAG~\citep{chang-etal-2025-main}.

\paragraph{Credit Assignment in Multi-Agent Reinforcement Learning.}
In multi-agent reinforcement learning (MARL), credit assignment under shared rewards is a well-studied problem. Methods such as QMIX~\citep{qmix} decompose the global value function into per-agent contributions, while counterfactual approaches like COMA~\citep{coma} estimate individual agent impact by comparing observed outcomes against baselines that marginalize out each agent's action. More recently, LLM-MCA~\citep{nagpal2025leveraging} utilizes LLM critics to decompose global rewards into individual feedback. However, it remains tied to updating the neural network weights of traditional reinforcement learning agents, and it relies on a task-specific attribution prompt. It adapts the comparative intuition behind counterfactual credit assignment to LLM-based workflows by comparing concrete alternative trajectories for the same query and using a prompted LLM to attribute observed differences in outcomes to individual agents. Unlike value-based decomposition methods such as QMIX, \alg does not impose a constraint requiring credits to reconstruct the global score. %

\paragraph{Prompt Optimization.}
Prompt design affects model behavior, with surface-level formatting choices alone producing substantial performance variation~\citep{sclar2024quantifying}.
EvoPrompt~\citep{guo-iclr24a} performs evolutionary search over prompt candidates using LLMs in its mutation and selection operators.
CAPO~\citep{zehle2025capo} builds on evolutionary search by incorporating AutoML techniques such as racing to reduce optimization cost, and introduces prompt length penalties.
TextGrad~\citep{yuksekgonul-nature25a} simulates gradient descent via LLM feedback, iteratively identifying and correcting failure cases in generated outputs.
MIPROv2~\citep{opsahl-ong-emnlp24a} samples prompt candidates and selects among them via Bayesian optimization, treating prompt search as a structured hyperparameter problem.
GEPA~\citep{agrawal2025gepareflectivepromptevolution} frames prompt optimization as a multi-objective problem over a Pareto front, where each dataset sample constitutes its own objective, and uses LLM-generated critiques of execution traces to guide prompt updates. These methods operate at the level of individual agents and can be applied to the \alg framework by utilizing them as local optimizers.

\paragraph{Optimization of Multi-Agent Systems.}
MIPROv2 in its multi-agent setting treats the joint prompt configuration as a single monolithic parameter, without decomposing the reward signal. GEPA approaches multi-agent configurations by updating agents sequentially via round-robin, optimizing one agent per step while holding all others fixed. Unlike these approaches, \alg{} decouples attribution from the update mechanism, enabling the use of arbitrary, sophisticated local optimizers. ADAS~\citep{hu2025automated} uses an optimizer LLM to iteratively propose and evaluate modifications to the MAS topology, while AFLOW~\citep{zhang2025aflow} formulates workflow optimization as Monte Carlo Tree Search over system configurations. MASS~\citep{zhou2025multi} interleaves prompt and topology optimization but propagates the global reward without agent-level attribution. More recent closed-source attribution-based works like HiveMind~\citep{xia2026hivemind} and MAPRO~\citep{zhang2026mapro} do isolate contributions, but couple their credit assignment to built-in LLM-oracle mutations. HiveMind relies on Shapley-based attribution, which requires per-agent credits to reconstruct the global score; \alg does not impose this decomposition. Furthermore, \alg{} drives node-level optimization through contrastive attribution grounded in reasoning traces across multiple joint rollouts, updates all agents simultaneously at each step, and leaves workflow topology fixed.

\section{Contrastive Attribution via In-Group Comparison}
\label{sec:method}

We formalize the optimization of a multi-agent system as a parameter optimization problem over joint agent configurations.
We first present the general problem setting, then introduce \alg{} and its contrastive attribution mechanism, and finally describe its instantiation for prompt optimization.

\subsection{Problem Formulation}
\label{sec:problem_formulation}

\paragraph{Agents.}
A multi-agent system (MAS) consists of $\Nagents$ agents $\{\agent{\agentidx}\}_{\agentidx=1}^{\Nagents}$.
Each agent $\agent{\agentidx}$ is a parameterized function mapping an input string to an observable output string,
\begin{equation}
    \agentout{\agentidx} = \agent{\agentidx}(\agentin{\agentidx};\, \agentparams{\agentidx}),
\end{equation}
where $\agentparams{\agentidx} \in \paramspace{\agentidx}$ denotes the local parameter set searched by a per-agent local optimizer $\optimizer{\agentidx}$.
The input $\agentin{\agentidx}$ may be a task query or the output of an upstream agent, which is injected into the agent's prompt.
Agent behavior is primarily governed by a prompt $\prompt{\agentidx}$ specifying role and instructions; additional local parameters could include, but are not limited to, model weights, model choice, or decoding settings.
Agents may interact with external tools, incorporating tool outputs into their computation, and may produce internal reasoning that is not part of the extracted output $\agentout{\agentidx}$.

\paragraph{Multi-Agent System.}
The MAS is a directed graph $\graph = (\nodes, \edges)$ with $|\nodes| = \Nagents$ agents as vertices.
Each edge $(u \rightarrow v) \in \edges$ carries the variables passed from agent $\agent{u}$ to agent $\agent{v}$, and the graph supports structured workflow designs including conditional routing, parallel execution, and ensembling.
A \textit{rollout} is a single execution of $\graph$ on a query $\queryidx$; the resulting \textit{trajectory} $\trajectoryplane$ is the ordered sequence of all agent outputs $\agentout{\agentidx}$ produced during that execution.
Given a query $\queryidx$ and a joint configuration $\jointconfig{} = \left(\agentparams{1}, \dots, \agentparams{\Nagents}\right)$, the graph produces a prediction $\pred = \graph(\queryidx;\, \jointconfig{})$.

\paragraph{Optimization Objective.}
Each task $\task$ is associated with a scoring function $\sysscore_\task(\pred)$ that maps a prediction to a global system score, e.g., a reward signal, human ranking, or accuracy in classification settings.
In this work, we focus on optimizing the local parameters $\jointconfig{}$, treating the workflow graph $(\nodes, \edges)$ as a fixed upstream design choice.
The objective is to find the joint configuration maximizing expected score over the data distribution $\querydist$:
\begin{equation}
    \argmax_{\jointconfig{}}
    \;\mathbb{E}_{\queryidx \sim \querydist}\!\left[
        \sysscore_\task\!\left(\graph(\queryidx;\, \jointconfig{})\right)
        \right].
        \label{eq:objective_infinite}
    \end{equation}
    In practice, optimization proceeds over a fixed query set $\queryset$:
\begin{equation}
    \argmax_{\jointconfig{}}
    \;\frac{1}{|\queryset|} \sum_{\queryidx \in \queryset}
    \sysscore_\task\!\left(\graph(\queryidx;\, \jointconfig{})\right).
    \label{eq:objective_finite}
\end{equation}

This optimization is subject to a finite budget, which can, for example, be defined in terms of tokens consumed by the downstream LLMs.

\paragraph{The Credit Assignment Problem.}
The reward signal $\sysscore_\task$ is defined at the system level, yet the parameters $\agentparams{\agentidx}$ that determine agent behavior are local.
Directly optimizing Equation~\ref{eq:objective_finite} over the joint space $\jointconfig{}$ is intractable for large $\Nagents$, as it requires exhaustive evaluation of numerous configurations.

\subsection{\alg}
\label{sec:cantante}

\begin{figure}
    \centering
    \includegraphics[width=0.85\linewidth, trim=0 4pt 0 5pt, clip]{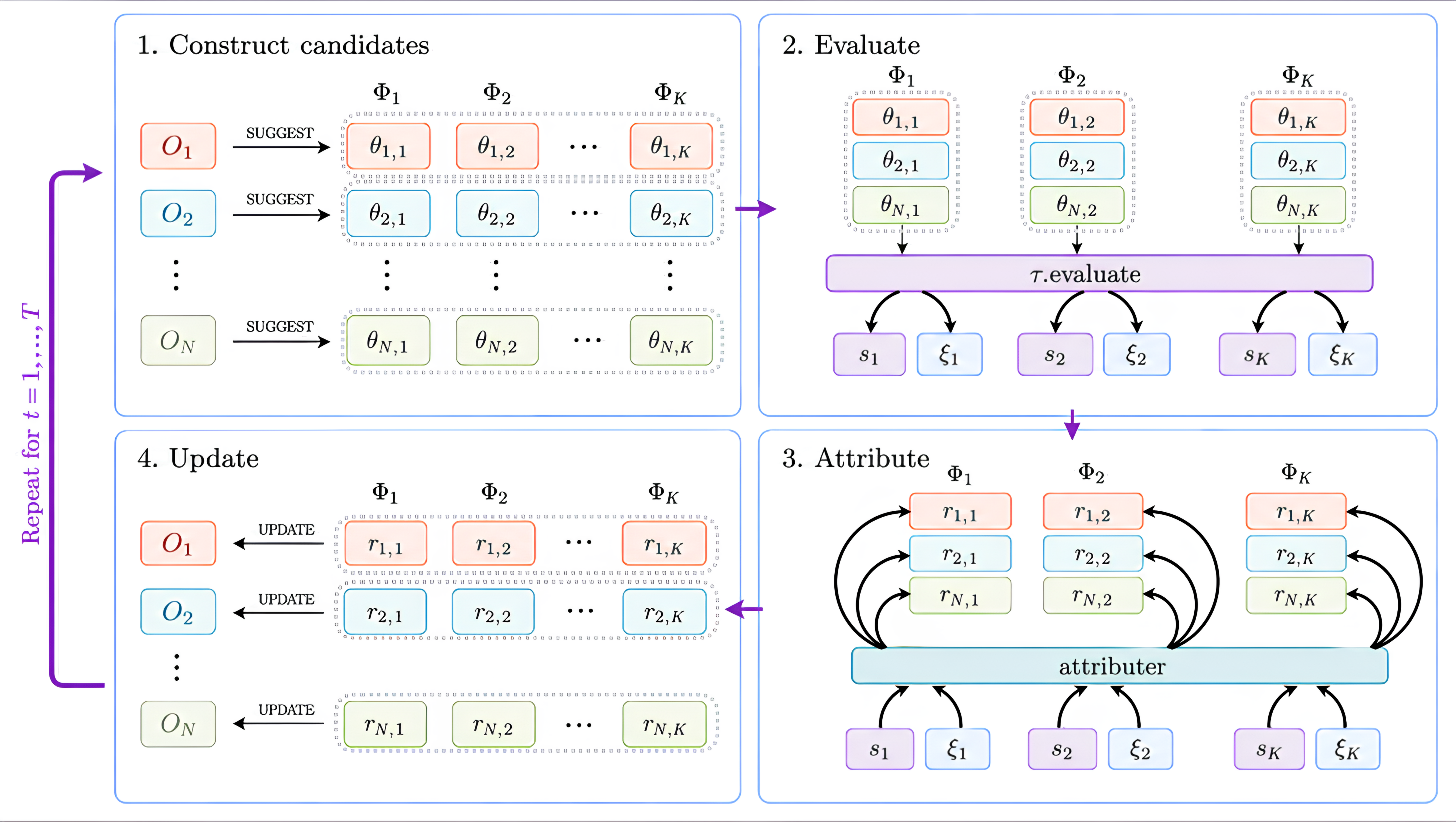}
    \caption{\textbf{Overview of \alg.} \textbf{(1)}~At each iteration, every local optimizer $O_a$ proposes $K$ candidate parameterizations $\theta_{a,i}$, yielding $K$ joint configurations. \textbf{(2)}~Each joint configuration is evaluated on task $\tau$, producing system-level scores $s_i$ and an execution trace $\xi_i$ per configuration. \textbf{(3)}~The attributer receives the full set of scores and traces and performs contrastive attribution within random comparison groups, estimating aggregated credit $r_{a,i}$. \textbf{(4)}~Each parameterization's credit is forwarded to its local optimizer $O_a$ to perform local update steps.}
    \label{fig:overview}
\end{figure}

\alg{} (\textbf{C}ontrastive \textbf{A}ttributio\textbf{n} for \textbf{T}uning of Multi-\textbf{A}ge\textbf{nt} Syst\textbf{e}ms) makes multi-agent systems learnable by decomposing the joint optimization over $\jointconfig{}$ into $\Nagents$ parallel local optimization problems, using contrastive attribution. We offer a visual overview in Figure~\ref{fig:overview}, and detail its pseudo-algorithm in Appendix~\ref{app:pseudo_alg}.

Assigning credit in absolute terms is difficult: judging whether an agent performed well requires knowledge of what constitutes good behavior for every task, which is not generally available. Relative attribution within a comparison group is more tractable: We utilize an attribution model that, for each agent, independently assesses how well its output contributed to the observed system outcome by comparing it with a group of contrastive rollouts.

This contrastive framing offers an additional advantage in degenerate cases where all configurations in a group achieve identical system scores. In these cases, a purely score-based optimizer would stall because the absence of score differences eliminates the signal to guide optimization. The attribution LLM, instead, can fall back on assessing intermediate agent outputs $\agentout{\agentidx}$ directly. As a result, \alg{} can still distinguish between configurations with identical system-level scores.

The attributer is a prompted LLM that receives a group of trajectories and their system scores and outputs a scalar credit estimate $\agentcredit{\agentidx} \in [-1, +1]$ per agent per rollout, where positive credit indicates that the agent's behavior was relatively supportive of high system performance and negative credit indicates the converse.
Concretely, the attributer receives the query $\queryidx$, the per-agent outputs $\{\agentout{\agentidx}\}_{\agentidx=1}^{\Nagents}$ extracted from each trajectory $\trajectory{\queryidx}{\configidx}$, and the system scores $\sysscore_\task(\pred)$. When an agent is invoked multiple times within a rollout, we use its final produced output as $\agentout{\agentidx}$. We assume the attributer to be instruction-following and non-adversarial, consistent with its prompt design, which includes a few-shot example grounding its behavior; the full prompt is provided in Appendix~\ref{app:attributer_prompt}.

At each iteration, the local optimizers propose $\Nconfigs$ candidate parameterizations per agent, which are assembled into $\Nconfigs$ joint configurations $\{\jointconfig{\configidx}\}_{\configidx=1}^{\Nconfigs}$ (Figure~\ref{fig:overview}, Panel 1). Exhaustively evaluating all $\Nconfigs^N$ cross-agent combinations would be prohibitively expensive. Instead, each joint configuration pairs one candidate from each agent, requiring only $\Nconfigs$ rollouts per query. The contrastive attributer then estimates contributions per agent parameterization despite the jointly varying co-parameterizations. The resulting joint configurations are evaluated on task $\tau$, producing system scores $s_i$, and execution traces $\xi_i$ (Figure~\ref{fig:overview}, Panel 2).

In the attribution stage (Figure \ref{fig:overview}, Panel 3), the attributer assesses each agent's contribution relative to the other configurations in its comparison group, yielding per-query credits $\credit{\queryidx}{\agentidx}{\configidx} \in [-1,+1]$. We decide to assign scalar credits rather than ranks, enabling more nuanced attribution. The signed range allows the attributer to express both positive and negative contributions while remaining within an interpretable, normalized range. We aggregate these credits into a per-agent-and-configuration score by averaging uniformly over the query set, mirroring the per-query weighting of the empirical objective in Equation~\ref{eq:objective_finite}: $ \aggcredit{\agentidx}{\configidx} = \frac{1}{|\queryset|} \sum_{\queryidx \in \queryset} \credit{\queryidx}{\agentidx}{\configidx}$.%
    
Finally, each local optimizer $\optimizer{\agentidx}$ consumes the resulting parameter-credit pairs $\{(\param{\agentidx}{\configidx}, \aggcredit{\agentidx}{\configidx})\}_{\configidx=1}^{\Nconfigs}$ (Figure \ref{fig:overview}, Panel 4) and proposes updated parameterizations for the next iteration.

\section{Experiments}
\label{sec:results}

We instantiate \alg for prompt optimization, treating agent prompts $\prompt{\agentidx}$ as learnable parameters and using CAPO as the default local optimizer $\optimizer{\agentidx}$. We evaluate \alg{} against GEPA and MIPROv2 on three benchmarks, spanning code generation (MBPP), mathematical reasoning (GSM8K), and multi-hop question answering (HotpotQA), each paired with a distinct workflow graph. We additionally analyze credit attribution behavior and conduct experiments to broaden understanding of the method.

\subsection{Setup}
\label{sec:exp}

\paragraph{Benchmarks and Graphs.}
We score MBPP~\citep{austin-arxiv21} as the proportion of programming tasks for which the system produces code that passes all test cases, using a conditional workflow with a planner, executor, and validator agent, with tool access for code execution.
GSM8K~\citep{cobbe-arxiv21a} is evaluated by exact match and uses an ensembling graph in which three parallel executor agents produce predictions that a consensus agent aggregates into a final answer.
HotpotQA~\citep{yang2018hotpotqa} is scored by exact match and uses an agentic retrieval-augmented generation graph with a retriever, reader, synthesizer, and hallucination checker. These benchmarks were chosen to cover three qualitatively different MAS structures: conditional execution, parallel ensembling, and retrieval-augmented multi-hop reasoning. Full experimental details, including implementation details, parameterizations, benchmark descriptions, as well as respective topologies, hardware, and model details, are provided in Appendix~\ref{app:exp_details}.

\paragraph{Models and Optimizers.}
All downstream task agents use Qwen3 with 30 billion total parameters, 3 billion of which are active. GPT-OSS-120B, as a model from a distinct family, serves as both an optimizer and an attribution model.
We evaluate \alg against GEPA~\citep{agrawal2025gepareflectivepromptevolution} and  MIPROv2~\citep{opsahl-ong-emnlp24a}, representing two dominant paradigms of MAS-capable prompt optimization.

\paragraph{Protocol.}
To ensure a fair starting point, all initial prompts are generated automatically by prompting GPT-OSS-120B, as detailed in Appendix~\ref{app:pc}. We report the initial prompts in Appendix~\ref{app:init_prompts}. All optimizers operate under a shared optimization budget of 10 million tokens, which includes input and output tokens from downstream task-agent calls, as well as optimizer and attribution calls.
The reported score is the development-set-selected configuration at the last optimization step completed within budget. On MBPP seeds 42 and 47, MIPROv2 did not return a configuration within the shared 10M-token budget. We extended its budget on these seeds to 15.8M and 11.8M tokens, respectively, favoring MIPROv2 in the comparison.
All results are reported as mean and standard deviation across three seeds using \textit{Bessel's correction}.

\subsection{Main Results}
\label{sec:main_results}

\begin{table}[t]
  \centering
  \caption{Test accuracy (\%) averaged over three seeds ($\pm$std). \textbf{Bold: best}, \underline{underlined: second-best} per benchmark. Average rank computed across benchmarks and seeds. The bottom rows show \alg's absolute gain over the initial prompt and the best baseline.}
  \label{tab:main-results}
  \begin{tabular}{llllc}
    \toprule
    \textbf{Optimizer} & \textbf{MBPP} & \textbf{GSM8K} & \textbf{HotpotQA} & \textbf{Avg Rank} \\
    \midrule
    Initial & \phantom{0}5.54\,$_{\pm\phantom{0}1.62}$ & 59.20\,$_{\pm10.73}$ & \phantom{0}9.67\,$_{\pm\phantom{0}3.00}$ & \phantom{-}3.44 \\
    GEPA & \underline{22.96}\,$_{\pm18.30}$ & 61.27\,$_{\pm\phantom{0}2.66}$ & 10.93\,$_{\pm\phantom{0}3.91}$ & \phantom{-}2.67 \\
    MIPROv2 & 18.42\,$_{\pm14.13}$ & \underline{69.80}\,$_{\pm\phantom{0}7.10}$ & \textbf{14.20}\,$_{\pm\phantom{0}5.72}$ & \phantom{-}\underline{2.33} \\
    \midrule
    \alg{} (ours) & \textbf{41.89}\,$_{\pm\phantom{0}7.56}$ & \textbf{82.33}\,$_{\pm\phantom{0}4.35}$ & \underline{11.93}\,$_{\pm\phantom{0}5.06}$ & \phantom{-}\textbf{1.44} \\
    $\Delta$ vs. Initial & \textcolor{cbGreen}{36.35} & \textcolor{cbGreen}{23.13} & \textcolor{cbGreen}{\phantom{-}2.27} & \textcolor{cbGreen}{-2.00} \\
    $\Delta$ vs. Best & \textcolor{cbGreen}{18.93} & \textcolor{cbGreen}{12.53} & \textcolor{cbOrange}{-2.27} & \textcolor{cbGreen}{-0.89} \\
    \bottomrule
  \end{tabular}
\end{table}

Table~\ref{tab:main-results} reports the main results.
Across the nine benchmark-seed combinations, \alg{} obtains the highest test score in 6 cases and achieves the best average rank of 1.44, compared to 2.33 for the strongest baseline MIPROv2.
\alg{} outperforms the unoptimized initial prompts on 8 of 9 seeds, with gains on MBPP and GSM8K that substantially exceed the cross-seed standard deviations. The highest-scoring prompt sets obtained by \alg{} are provided in Appendix~\ref{app:best_prompts}.

On MBPP, \alg{} achieves a mean accuracy of \SI{41.89}{\%}, improving over the strongest baseline by 18.93 percentage points (pp) and over the unoptimized initial prompts by \SI{36.35}{\pp}.
Beyond mean performance, \alg{} exhibits substantially lower standard deviation than the baselines ($\pm$~\num{7.56} vs.\ $\pm$~14--18 pp). The relatively low variance across seeds indicates more stable optimization behavior, suggesting that per-agent attribution provides a more consistent update signal than system-level feedback alone.

On GSM8K, \alg achieves \SI{82.33}{\%}, improving over MIPROv2 by \SI{12.53}{\pp} and over GEPA by  \SI{21.06}{\pp}. Its variance is lower than MIPROv2's but higher than GEPA's ($\pm$~\num{4.35} vs. $\pm$~\num{7.10} and $\pm$~\num{2.66} pp, respectively). The high variance of the initial prompts ($\pm$ \num{10.73} pp) indicates that GSM8K performance is particularly sensitive to prompt quality. Notably, GEPA fails to meaningfully improve over the initial prompts, plateauing near the baseline performance.

On HotpotQA, performance differences across all optimizers and for unoptimized prompts are small (ranging from \SI{9.67}{\%}  to  \SI{14.2}{\%}) relative to cross-seed standard deviations (ranging from \SI{3}{\%} to \SI{5.72}{\%}), suggesting that prompt optimization has limited leverage on this benchmark regardless of the method. Within this constrained regime, \alg remains within one standard deviation of the strongest baseline and improves over the unoptimized prompts.

\begin{figure}[t]
    \centering
    \includegraphics[width=0.63\textwidth]{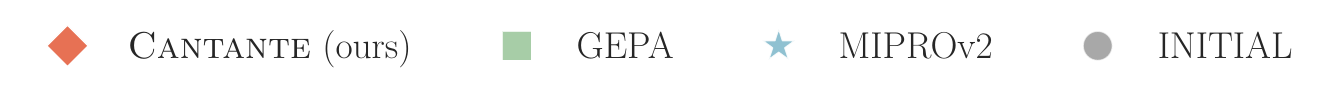}
    \\[4pt]

    \begin{minipage}[b]{0.32\textwidth}
        \centering
        \includegraphics[width=\textwidth]{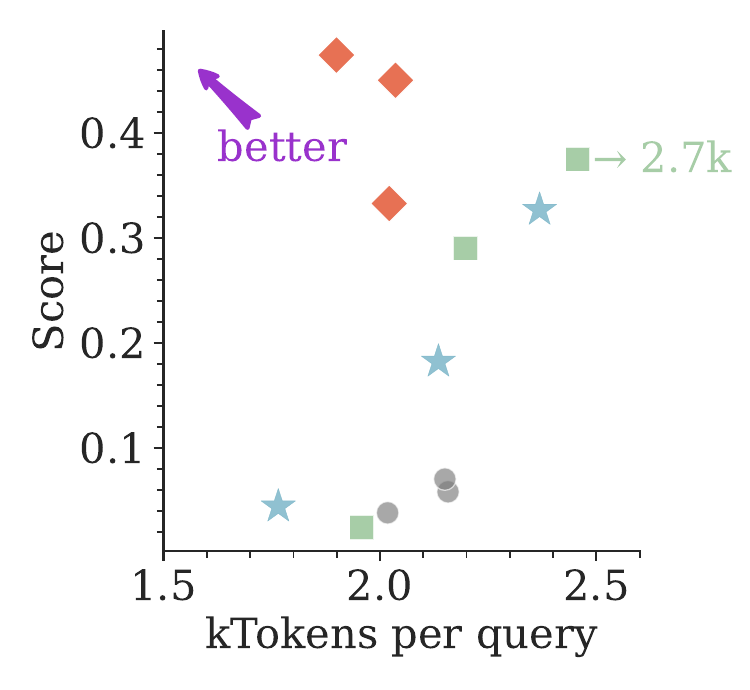}
        \subcaption{MBPP}
        \label{fig:pareto_mbpp}
    \end{minipage}
    \hfill
    \begin{minipage}[b]{0.32\textwidth}
        \centering
        \includegraphics[width=\textwidth]{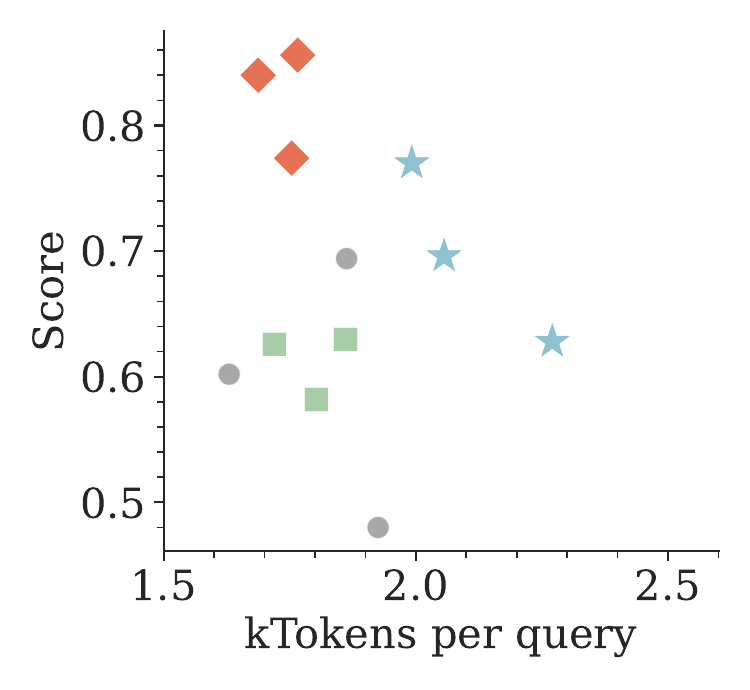}
        \subcaption{GSM8K}
        \label{fig:pareto_gsm8k}
    \end{minipage}
    \hfill
    \begin{minipage}[b]{0.32\textwidth}
        \centering
        \includegraphics[width=\textwidth]{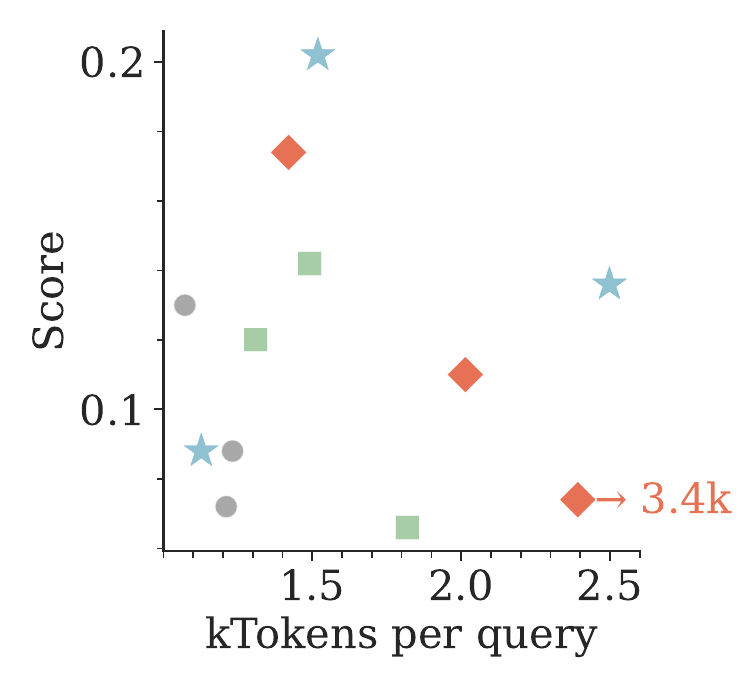}
        \subcaption{HotpotQA}
        \label{fig:pareto_hotpotqa}
    \end{minipage}
    \caption{Accuracy vs. evaluation-time inference cost across benchmarks, where inference cost denotes mean token usage per query for the final selected prompt sets.}
    \vspace*{-2em}

    \label{fig:pareto}
\end{figure}

Figure~\ref{fig:pareto} compares evaluated optimizers in terms of inference cost and achieved score. We distinguish the optimization budget from the evaluation-time inference cost. The former counts all input and output tokens consumed during optimization, including those consumed by downstream, optimizer, and attribution calls. The latter measures the tokens consumed (both input and output) by the final selected MAS when evaluated on the held-out test set, averaged over queries.
On MBPP and GSM8K, \alg{} achieves the strongest accuracy while simultaneously requiring the fewest mean evaluation-time tokens per inference (1.99 and 1.74 thousand tokens per invocation), beating the token efficiency of the unoptimized initial prompts (2.11 and 1.81), as well as MIPROv2 (2.09 and 2.11) and GEPA (2.28 and 1.79), demonstrating that the accuracy gains on these datasets are not an artifact of expensive reasoning traces, but rather a direct result of role-specific parameter updates.
In contrast, on the retrieval-augmented HotpotQA benchmark, \alg{} favors a more expansive configuration, using substantially more tokens per inference (2.28 vs. 1.17 for the initial prompts, 1.71 for MIPROv2, and 1.54 for GEPA). This increased cost reflects a regime where the optimizer attempts to resolve complex multi-hop reasoning through more detailed prompting, yet remains constrained by underlying topological bottlenecks that limit further performance or efficiency gains. A detailed breakdown of token usage across optimizers and benchmarks is provided in Appendix~\ref{app:tokens}.

\subsection{Analysis of Attributed Credits} \label{sec:credit_analysis}
\begin{wrapfigure}{r}{0.5\linewidth}
    \vspace*{-1em}
    \centering
    \includegraphics[width=\linewidth]{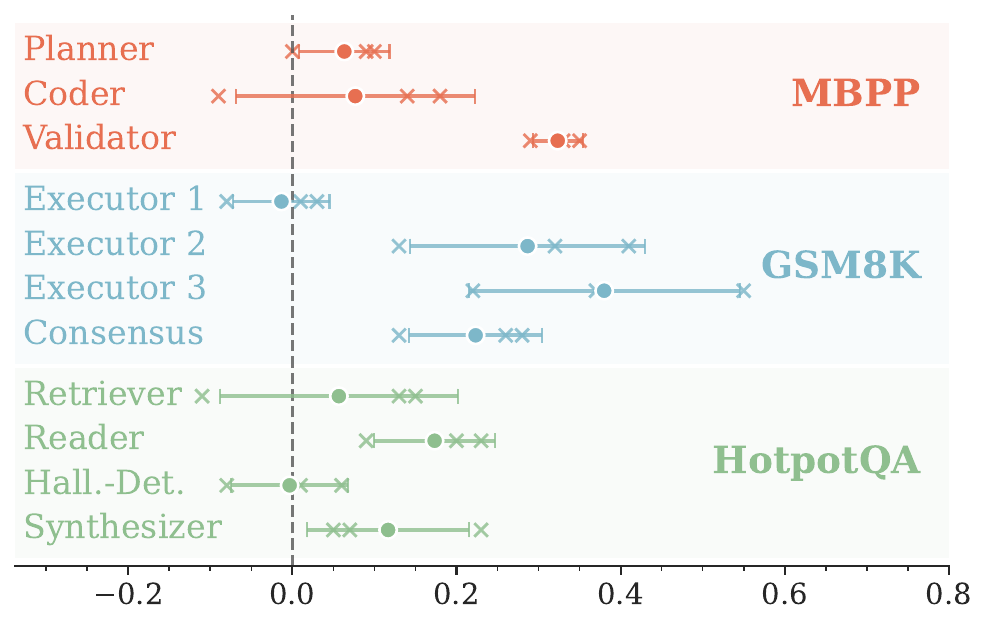}
    \caption{Spearman correlation between attribution credits and system scores, per agent and benchmark. Dots denote mean; crosses per-seed values; whiskers standard deviation.}
    \label{fig:credit_corr_mean}
    \vspace*{-2em}
\end{wrapfigure}

To further analyze our method's behavior, we examine the credits assigned in the main experiments by computing the Spearman correlation $\rho$ between the system score for a joint configuration and the attributions to the respective agent prompts received during optimization. Figure~\ref{fig:credit_corr_mean} reports this analysis for all three benchmarks, and Appendix~\ref{app:credit-correlation} provides the per-seed values.

On MBPP, the validator receives the highest and most stable attribution across seeds, suggesting that it is a particularly relevant target for optimization. The planner and coder show substantially weaker correlations. This suggests that, within this topology, many prompt variants are sufficient for these roles, or the validator's final judgment constrains their outputs. %

On GSM8K, an ordering emerges across the three executor agents despite their identical tasks: Executor~1 receives near-zero attribution, while Executors~2 and~3 carry progressively stronger signals. This pattern is consistent with recency bias documented by \citep{liu-etal-2024-lost} and \citep{zehle2026can}: because executor outputs appear in a fixed order in the consensus prompt, later outputs, especially that of Executor~3, are more likely to be attended to and therefore receive stronger attribution. The consensus agent receives moderate attribution, as expected from its role in aggregating the executor outputs.

On HotpotQA, the hallucination detector receives near-zero attribution across seeds, and the retriever straddles zero, while the reader and synthesizer show moderate positive signal. Although these magnitudes are not far below some agents in MBPP and GSM8K, the pattern is less sharply differentiated: no single agent dominates the correlation analysis. This may indicate that structural bottlenecks limit the extent to which improvements in any individual prompt affect the final system score. Alternatively, it may reflect a limitation of the attributer, which may have struggled to identify meaningful per-agent contributions in this setting.

\subsection{Additional Experiments}
\label{sec:ablations}

\begin{wraptable}{r}{0.4\textwidth}
    \vspace*{-3em}
  \centering
  \caption{Ablation results on GSM8K (Seed 42). Test accuracy relative to the default setting, in percentage points.}
  \label{tab:ablations}
  \begin{tabular}{lr}
    \toprule
    \textbf{Configuration} & \textbf{$\Delta$ Score} \\
    \midrule
    Default setting & 84.00 \\
    \midrule
    \textit{Identity Attribution} &  \\
    \hspace{1em}Equal Steps & $-$\,13.40 \\
    \hspace{1em}Equal Budget & $-$\,\phantom{0}4.40 \\
    \midrule
    \textit{Attribution component} &  \\
    \hspace{1em}alt. Prompt & $+$\,\phantom{0}0.60 \\
    \hspace{1em}alt. Model & $+$\,\phantom{0}0.20 \\
    \midrule
    \textit{Prompt optimizer} &  \\
    \hspace{1em}w/ Few-Shots & $-$\,\phantom{0}9.20 \\
    \hspace{1em}alt. Optimizer & $+$\,\phantom{0}3.20 \\
    \midrule
    \textit{Group size} &  \\
    \hspace{1em}g = 2 & $-$\,54.40 \\
    \hspace{1em}g = 5 & $+$\,\phantom{0}0.80 \\
    \midrule
    \textit{Dataset size} &  \\
    \hspace{1em}|DS| = \phantom{0}75 & $-$\,\phantom{0}2.80 \\
    \hspace{1em}|DS| = 150 & $-$\,\phantom{0}3.40 \\
    \bottomrule
  \end{tabular}
  \vspace*{-2em}

\end{wraptable}

We analyze the sensitivity of our method to key design choices and hyperparameters by conducting additional experiments on GSM8K, ablating the attributer components, alternating the local optimizer, and testing sensitivity to the group-size parameter and to varying dataset sizes. Table~\ref{tab:ablations} reports the performance differences against the default setting from the main experiments.

To isolate the impact of the attributer, we replaced it with an ``identity attributer'' that propagates the global system score directly to the local optimizer. Under equal optimization steps, this removal causes a severe \SI[round-precision=1]{13.4}{\pp} drop in accuracy, highlighting contrastive attribution as the primary driver of optimization efficiency. Because the identity attributer bypasses LLM attribution costs, we also evaluated an ``equal budget'' setting that allows the local optimizer to execute substantially more optimization steps. Given this extended budget, the identity attributer alone achieves \SI[round-precision=1]{79.60}{\%}, which is roughly \SI[round-precision=1]{9.8}{\pp} higher than the strongest baseline (MIPROv2). Yet, default \alg{} still outperforms this budget-advantaged run by \SI[round-precision=1]{4.4}{\pp}. This breakdown clarifies that while a strong local optimizer sets a high performance floor, per-agent attribution yields a superior learning signal than global rewards.

Exchanging the attributer prompt or the LLM underlying the optimization and attribution model  (Qwen3-Next-80B instead of GPT-OSS-120B) yields scores of \SI[round-precision=1]{84.60}{\%} and \SI[round-precision=1]{84.20}{\%} against the default of \SI[round-precision=1]{84.00}{\%}, which is within the cross-seed standard deviation of the original experiment. This suggests that the method is robust against concrete choices of prompt and LLMs.

Prior work suggests performance gains from introducing few-shot examples into prompt optimization \citep{zehle2025capo, wan2024teach}. However, bootstrapping these from initial rollouts caused a \SI[round-precision=1]{9.2}{\pp} drop in test accuracy. This performance decay suggests that, when the initial agent configuration is weak, its rollouts provide poor demonstrations of the intended agent behavior—reusing these traces as exemplars biases later prompt candidates toward the same flawed reasoning patterns. The credit correlation analysis in Appendix~\ref{app:credit-correlation} further shows that introducing few-shot examples causes the attributer to assign elevated and less differentiated credit-score correlations across agents, suggesting that the resulting agent behaviors become less agent-specific.

To test the generalizability of \alg, we initialized it with EvoPrompt as a local optimizer and observed a \SI[round-precision=1]{3.2}{\pp} increase in performance. This suggests that our method successfully extends to other local optimizers and that performance gains cannot be attributed solely to CAPO.

To understand the impact of the group-size hyperparameter $\groupsize$, which controls the number of configurations within a contrastive comparison group, we vary it around the default $\groupsize=3$.\footnote{We additionally increase the number of parameterizations per agent from 9 to 10 to ensure divisibility between $\Nconfigs$ and $\groupsize$.} At $\groupsize=2$, each group contains only a single pair of configurations. This binary contrast yields a severe performance drop of \SI[round-precision=1]{54.4}{\pp}. The credit correlation analysis in Appendix~\ref{app:credit-correlation} shows that the aggregate attribution direction is largely preserved at $\groupsize=2$, indicating the collapse is driven by elevated per-query credit variance rather than systematic misdirection — a single pair of rollouts might lack the overlapping variation needed to disentangle individual agent contributions reliably. The \SI[round-precision=1]{0.8}{\pp} gain from $\groupsize=5$ over the default indicates diminishing returns for larger group sizes, supporting the premise that $\groupsize=3$ provides an effective trade-off between attribution accuracy and sample efficiency.

Finally, even when reducing the development set by half or three-quarters, \alg{} retains most of its performance. With 150 and 75 samples, it achieves \SI[round-precision=1]{80.60}{\%} and \SI[round-precision=1]{81.20}{\%}, corresponding to modest drops of only \SI[round-precision=1]{3.40}{\pp} and \SI[round-precision=1]{2.80}{\pp} relative to the default \SI[round-precision=1]{84.00}{\%} setting. This suggests that \alg{} extracts useful credit signals from a few examples and remains effective in low-data regimes.

\subsection{Discussion}
\label{sec:discussion}
\alg{} achieves the best average rank across benchmarks, outperforming the strongest baseline by up to \SI[round-precision=1]{18.9}{\pp}; on MBPP and GSM8K, these gains are obtained while improving over the token efficiency of the initial prompts, showing that improvements are not driven by increased test-time compute. 

The credit correlation analysis shows that the attributer does not merely echo system scores. These patterns can help diagnose MAS topology by indicating which agents are associated with performance gains. On HotpotQA, weaker correlations suggest that prompt optimization has limited leverage, due to topological weaknesses of the respective MAS.

The identity attributer ablation further shows that global-score propagation remains at a comparable level to MIPROv2 under an equal-steps setting, but remains \SI[round-precision=1]{13.4}{\pp} below \alg{}. Together with the robustness to the attributer prompt and model choice, this suggests that the central mechanism driving performance is the in-group comparison structure. We also find that \alg{} generalizes to other local optimizers and remains effective under substantially reduced development sets.

\section{Limitations and Future Work}
\label{sec:future_work}
\paragraph{Limitations.}
\alg{} holds the workflow graph topology fixed throughout optimization, treating the number of agents, edge structure, and tool assignments as upstream design choices. Performance, therefore, depends on the quality of this initial graph design.
The attribution model is a prompted LLM; consequently, attribution quality is bounded by the capabilities of the underlying model. In settings where the attributer is weaker than the downstream agents, credit estimates may become unreliable.
The use of an LLM as an attributer model incurs additional optimization budget, which might be allocated to additional optimization steps.
Finally, as the number of agents grows, the attributer receives increasingly long trajectories, which may exceed context limits or degrade attribution quality due to information overload. Our experiments cover topologies of up to four agents; scalability to substantially larger systems remains empirically untested. Many practical agentic workflows, however, use a moderate number of agents, and \alg remains directly applicable in this regime.

\paragraph{Future Work.}
The most natural extension is to lift the fixed-graph assumption, using the credit correlations analyzed in Section~\ref{sec:credit_analysis} as a signal for topology optimization: agents whose credits show consistently low correlation with system performance may indicate structural bottlenecks warranting redesign. Extending the local optimizer interface to model weights or decoding configurations would allow \alg{} to operate as a general MAS training framework beyond prompt optimization. A further direction is to revisit few-shot prompt optimization under warm-started settings, where demonstrations are bootstrapped only after the system has reached a sufficiently reliable initial configuration. Finally, incorporating richer supervision sources, such as execution traces or human preference signals, can meaningfully guide attribution quality.

\section{Conclusion}
\label{sec:conclusion}
We argued that optimizing LLM-based multi-agent systems is fundamentally a credit-assignment problem, a challenge well recognized in MARL but underexplored in  MAS. We introduced \alg{}, a contrastive attribution framework that converts system-level reward into per-agent update signals. \alg{} achieves the best average rank and, on two of three benchmarks, while matching or improving the token efficiency of the unoptimized prompts. The credit correlation analysis suggests that the attributer produces meaningful per-agent signals rather than echoing global scores. We see this work as evidence that contrastive attribution is a more effective optimization primitive for LLM-based MAS than global score propagation, and a step toward treating multi-agent system parameters as objects of machine learning rather than manual configuration.

\paragraph{Broader Impact.}
\alg{} reduces the manual effort required to configure MAS, improving efficiency and lowering deployment barriers. As a method for improving MAS, it inherits its dual-use risks, including misuse for disinformation or harmful automation, but does not differentially enable such applications beyond the underlying models.

\newpage

\section*{Acknowledgments}
Tom Zehle received funding by the European Union. This work was supported by the European Union's Horizon Europe research and innovation program under grant agreement No. 101214398 (ELLIOT).

\bibliographystyle{unsrtnat}
\bibliography{bibliography/strings, bibliography/references, bibliography/bib_automl, bibliography/bib_prompt_tuning, bibliography/proc, bibliography/my_proc}
\clearpage

\appendix
\section{Extended Results}
\subsection{Token Breakdown}\label{app:tokens}
We provide a detailed breakdown of the inference cost for the optimized prompts, in thousands of tokens per invocation. Table~\ref{tab:token-usage} reports the token usage of the optimizers on the main experiments in Section~\ref{sec:exp}, aggregated over seeds with mean and standard deviation, using Bessel's correction.

Table~\ref{tab:frac-meta-tokens} reports the fraction of token budget utilized by optimization and attribution LLMs per optimization method, for the main experiments in Section~\ref{sec:exp}.

Table~\ref{tab:ablation-tokens} reports the token usage for the further experiments in Section~\ref{sec:ablations}.
\begin{table}[h]
  \centering
  \caption{Evaluation-time token usage on the held-out test sets. We report the mean number of input and output tokens per inference, in thousands, with standard deviation across three seeds using Bessel's correction. \textbf{Bold} and \underline{underlined values} indicate the \textbf{lowest} and \underline{second-lowest} mean token usage for each benchmark, respectively.}
  \label{tab:token-usage}
  \begin{tabular}{lcccc}
    \toprule
    \textbf{Optimizer} & \textbf{MBPP} & \textbf{GSM8K} & \textbf{HotpotQA} & \textbf{Avg Rank} \\
    \midrule
    Initial & \phantom{0}2.11\,$_{\pm\phantom{0}0.08}$ & \phantom{0}1.81\,$_{\pm\phantom{0}0.16}$ & \phantom{0}\textbf{1.17}\,$_{\pm\phantom{0}0.09}$ & \textbf{2.11} \\
    GEPA & \phantom{0}2.28\,$_{\pm\phantom{0}0.36}$ & \phantom{0}\underline{1.79}\,$_{\pm\phantom{0}0.07}$ & \phantom{0}\underline{1.54}\,$_{\pm\phantom{0}0.26}$ & 2.78 \\
    MIPROv2 & \phantom{0}\underline{2.09}\,$_{\pm\phantom{0}0.30}$ & \phantom{0}2.11\,$_{\pm\phantom{0}0.15}$ & \phantom{0}1.71\,$_{\pm\phantom{0}0.71}$ & 3.00 \\
    \midrule
    \alg{} (ours) & \phantom{0}\textbf{1.99}\,$_{\pm\phantom{0}0.07}$ & \phantom{0}\textbf{1.74}\,$_{\pm\phantom{0}0.04}$ & \phantom{0}2.28\,$_{\pm\phantom{0}1.02}$ & \underline{2.11} \\
    \bottomrule
  \end{tabular}%
\end{table}

\begin{table}[h]
  \centering
  \caption{Fraction of optimization tokens spent on meta (attribution and optimization) calls, as a percentage of total tokens used up to the token budget. Values are averaged across seeds.}
  \label{tab:frac-meta-tokens}
  \begin{tabular}{lccc}
    \toprule
    \textbf{Optimizer} & \textbf{MBPP} & \textbf{GSM8K} & \textbf{HotpotQA} \\
    \midrule
    GEPA & 24.74\,$_{\pm\phantom{0}1.56}$ & 24.19\,$_{\pm\phantom{0}5.17}$ & 23.41\,$_{\pm\phantom{0}2.15}$ \\
    MIPROv2 & \phantom{0}0.45\,$_{\pm\phantom{0}0.18}$ & \phantom{0}0.82\,$_{\pm\phantom{0}0.06}$ & \phantom{0}0.66\,$_{\pm\phantom{0}0.09}$ \\
    \alg{} (ours) & 25.91\,$_{\pm\phantom{0}1.49}$ & 25.55\,$_{\pm\phantom{0}1.48}$ & 20.11\,$_{\pm\phantom{0}2.25}$ \\
    \bottomrule
  \end{tabular}
\end{table}

\begin{table}[h]
  \centering
  \caption{Evaluation token usage per MAS invocation for ablation configurations on GSM8K, seed 42. Values are average evaluation tokens divided per query, in thousands.}
  \label{tab:ablation-tokens}
  \begin{tabular}{lc}
    \toprule
    \textbf{Configuration} & \textbf{Tokens/inv (k)} \\
    \midrule
    Default setting & \phantom{0}1.69 \\
    \midrule
    Attribution Prompt & \phantom{0}1.63 \\
    Attribution Model & \phantom{0}1.97 \\
    w/ Few-Shot Examples & \phantom{0}2.44 \\
    |DS| = 75 & \phantom{0}1.72 \\
    |DS| = 150 & \phantom{0}1.66 \\
    |G| = 2 & \phantom{0}1.96 \\
    |G| = 5 & \phantom{0}1.65 \\
    Identity Attribution (Equal Budget) & \phantom{0}1.66 \\
    Identity Attribution (Equal Steps) & \phantom{0}1.66 \\
    PO = EvoPrompt & \phantom{0}1.83 \\
    \bottomrule
  \end{tabular}
\end{table}

\newpage

\subsection{Further Analysis of Credit Attribution} \label{app:credit-correlation}
\begin{figure}[h]
    \centering
    \subcaptionbox{MBPP\label{fig:credit_corr_mbpp}}{%
        \includegraphics[width=0.32\linewidth]{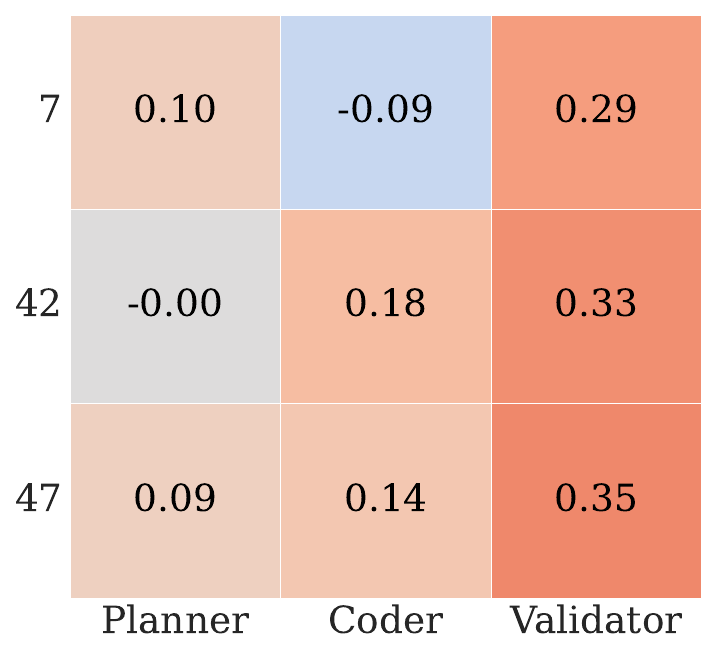}}%
    \hfill
    \subcaptionbox{GSM8K\label{fig:credit_corr_gsm8k}}{%
        \includegraphics[width=0.32\linewidth]{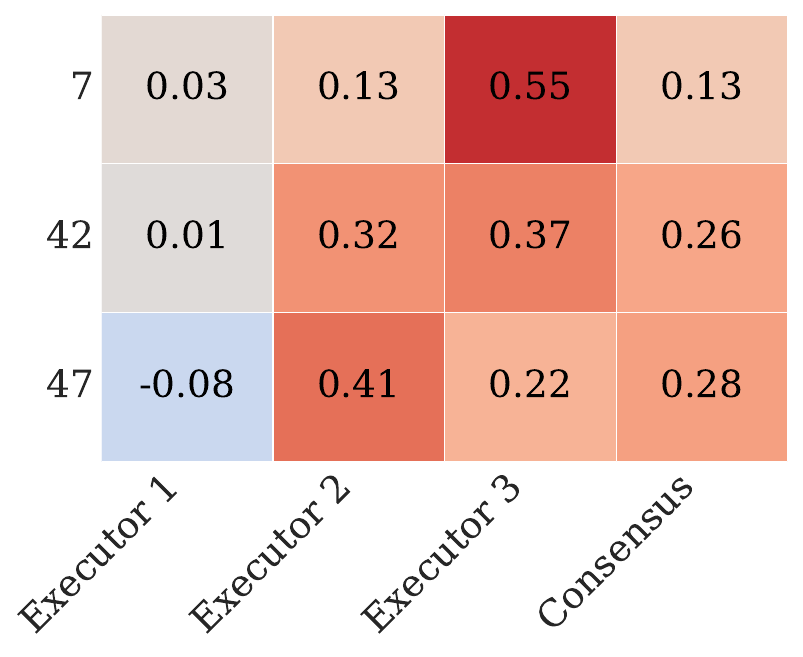}}%
    \hfill
    \subcaptionbox{HotpotQA\label{fig:credit_corr_hotpotqa}}{%
     \includegraphics[width=0.32\linewidth]{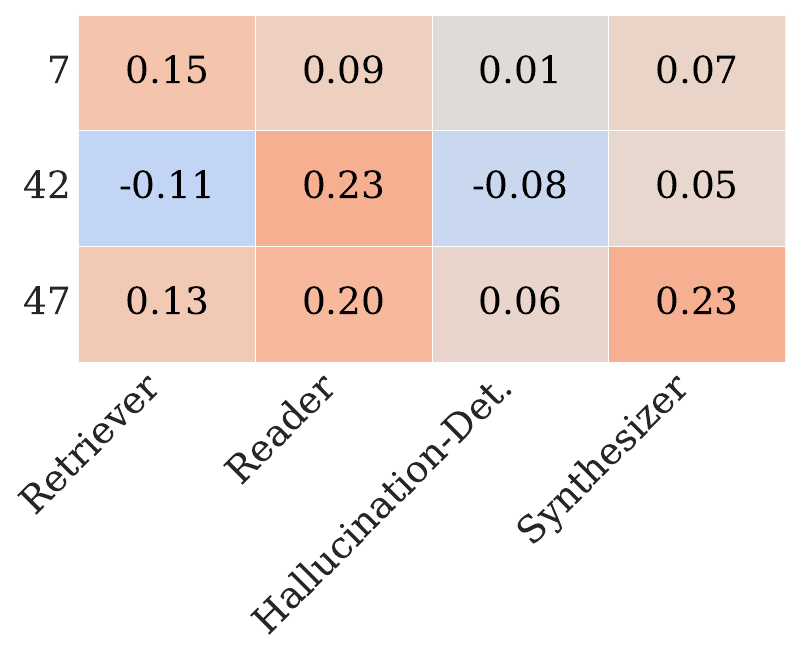}}
    \caption{Spearman correlation ($\rho$) between attribution credits and system scores, per agent and seed.}
    \label{fig:credit_corr_gsm8k_hotpotqa}
\end{figure}

Figure~\ref{fig:credit_corr_gsm8k_hotpotqa} provides the per-seed Spearman correlations underlying the aggregated analysis in Section~\ref{sec:credit_analysis}.

\begin{wrapfigure}{r}{0.5\linewidth}
    \vspace*{-2em}
    \centering
    \includegraphics[width=\linewidth]{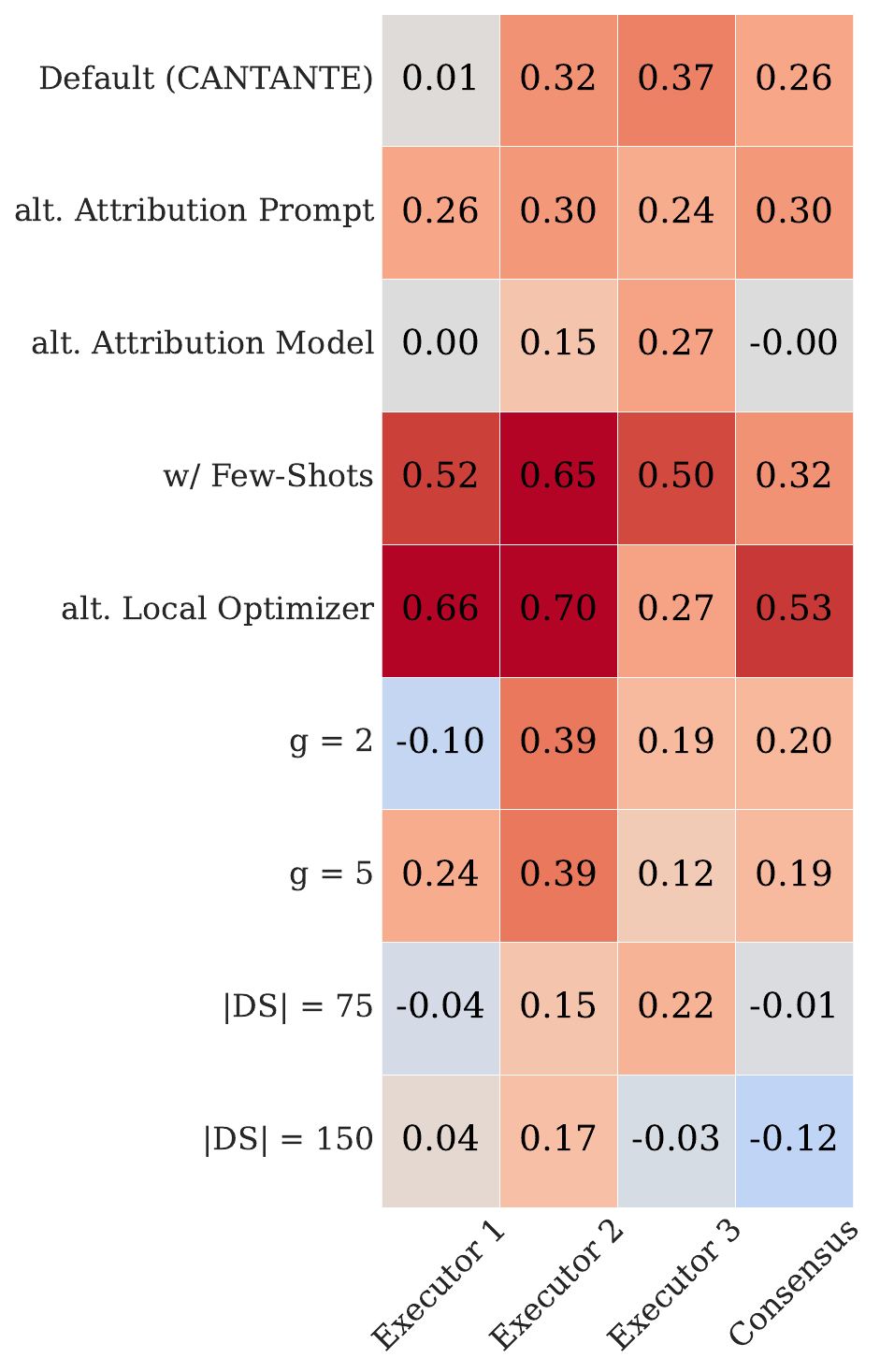}
    \caption{Spearman correlation between attribution credits and system scores for the additional experiments, on GSM8K, seed 42.}
    \label{fig:credit_corr_ablations}
    \vspace*{-2em}
\end{wrapfigure}

Figure~\ref{fig:credit_corr_ablations} extends the credit correlation analysis of Section~\ref{sec:ablations} to the ablation configurations. Across most settings -- including an alternative attribution prompt, alternative attribution model, and varying dataset sizes -- the correlation structure remains consistent with the default: Executor 1 receives near-zero credit correlation while Executors 2 and 3 carry the dominant signal, reflecting the positional bias discussed in Section 4.3. This stability supports the robustness of the attribution mechanism to concrete design choices.
The most notable deviation occurs in the w/ Few-Shots configuration, where all four agents show elevated and roughly uniform correlations (\num{0.32}-\num{0.65}). This flattening of the per-agent signal indicates that the attributer reverts to echoing the global system score rather than isolating individual contributions — providing a potential explanation for the \SI{9.2}{\pp} accuracy drop reported in Table~\ref{tab:ablations}.
For $\groupsize=2$, the aggregate correlation structure is largely preserved relative to the default, with Executor 1 remaining near zero. This suggests the severe performance degradation at $\groupsize=2$ is not caused by systematic misdirection of credits, but is instead consistent with elevated per-query credit variance. With only two rollouts available for comparison, individual credit estimates become unreliable even when their average direction is maintained.

\section{Experimental Details} \label{app:exp_details}
\subsection{Implementation Details}
We implement CAPO and EvoPrompt using promptolution \citep{zehle2026promptolution} and MIPROv2 and GEPA using DSPy \citep{khattab-iclr24a}. Agent architectures are constructed with LangGraph \citep{langgraph2024}. Experiments are conducted on NVIDIA L40 GPUs. The main experiments took roughly 100 GPU-hours to finish, and the ablation studies 30, totaling 130 GPU-hours.

\subsection{Parameterizations}
We use the default parameters of the respective implementations where possible. To control for computational budget, we apply the following modifications. For MIPROv2, we reduce the number of trials to 10 and set the number of candidates to 4. For GEPA, we set the maximum number of metric calls to 150 as suggested in the documentation, with optimization allowed to continue in increments of 150 per step. For CAPO and EvoPrompt, we reduce the population size to 6 and the number of offspring to 3; additionally, for CAPO, we reduce the block size to 10 and the maximum number of block evaluations to 5. In the group-size ablation, the number of offspring is increased to 4 to ensure the divisibility of the group and configuration sizes.
For LLM inference, we use a temperature of 0.7 for the downstream model and 0.1 for the optimizer LLM. Maximum token lengths for the downstream model are set to 512 for GSM8K and HotpotQA, and 1024 for MBPP. The optimizer model operates with a context window of 4096 tokens, extended to 8192 in the group size ablation to accommodate five rollouts within the attributer's context.

\subsection{Benchmark Details} \label{app:benchmark} 

\subsubsection{Seeding}
For each benchmark, we sample a development set of 300 instances and an evaluation set of 500 instances from the respective official splits. The random seed controls LLM generation, the train-test split sampling, the sampling of initial prompts for the optimizer, and the random partitioning of agents into groups within \alg. We report results across three seeds (7, 42, 47). Note that LLM inference is not fully replicable due to hardware-level non-determinism.

\subsubsection{Scoring and Topologies}
\paragraph{MBPP}\citep{austin-arxiv21} is a code generation benchmark consisting of crowdsourced Python programming tasks, each accompanied by a natural-language description and a set of unit tests.
Following the standard evaluation protocol, a task is scored as 1 if the system produces code that passes all hidden test cases upon termination of the multi-agent loop, and 0 otherwise; the reported metric is the fraction of tasks solved on the held-out test split.
During execution, the validator agent has access to one of the three test cases per query; the remaining two are held out for scoring and are not observed by any agent at any point during the rollout.

The workflow graph consists of three agents arranged in a conditional loop.
A \textit{planner} agent receives the natural language task description and produces a step-by-step implementation plan.
The \textit{executor} agent has access to a code execution tool that enables it to write, run, and modify Python code within the rollout.
The \textit{validator} agent evaluates the generated code against the visible test case and, when needed, returns targeted feedback to the executor for repair.
If no bug is detected or the maximum number of repair iterations is reached, the loop terminates, and the final code is submitted for scoring; we set this maximum to 3.

\paragraph{GSM8K}~\citep{cobbe-arxiv21a} is a benchmark of grade-school mathematical reasoning problems requiring multi-step arithmetic and algebraic reasoning, scored by exact match against the ground-truth numerical answer.

The workflow graph implements a parallel ensembling strategy.
Three \textit{executor} agents independently receive the query, and each produces a numerical prediction together with a step-by-step justification.
A \textit{consensus} agent receives all three predictions and justifications and produces the final answer by aggregating the evidence.
No agent in this graph has access to external tools; all computation is performed via language model inference.

\paragraph{HotpotQA}~\citep{yang2018hotpotqa} is a multi-hop question-answering benchmark that requires synthesizing information from multiple supporting documents and is scored by exact match against the ground-truth answer string.
The benchmark is designed to probe robustness to distractor passages and to test multi-step reasoning over retrieved evidence.

The workflow graph implements an agentic retrieval-augmented generation pipeline.
A \textit{retriever} agent receives the query and issues targeted queries to a document store via a retrieval tool, returning a set of relevant passages.
A \textit{reader} agent receives the retrieved passages and extracts the facts most relevant to the query.
A \textit{synthesizer} agent receives the extracted facts and produces a candidate answer.
A \textit{hallucination checker} agent receives the candidate answer together with the supporting passages and verifies factual consistency; if a hallucination is detected, it generates corrective feedback and conditionally routes execution back to the reader agent, forming a verification loop.
If no hallucination is detected or if the maximum number of verification iterations is reached, the candidate answer is submitted for scoring; we set this maximum to 3.
The retriever agent has access to a document retrieval tool; all other agents operate via language model inference only.

\subsection{Model Details}
We evaluate all optimizers using three language models: Qwen3-30B-A3B, Qwen3-Next-80B-A3B \citep{qwen3technicalreport}, and GPT-OSS-120B \citep{openai2025gptoss120b}. All three are mixture-of-experts models and are used as released, without additional quantization. Models can reason within the agentic loop, as the workflow requires them to produce structured outputs with explicit output tags, yielding their respective intermediate outputs $\agentout{\agentidx}$. 
All models are served via an API of a compute cluster using vLLM version 0.17~\citep{kwon2023vllm}.

\section{Pseudo Algorithm of \alg} \label{app:pseudo_alg}
Algorithm~\ref{alg:cantante} provides the pseudo code of \alg, summarizing the method described in Section~\ref{sec:method}.
\begin{algorithm}[h]
\caption{\alg (Contrastive Attribution for Tuning of Multi-Agent Systems)}
\label{alg:cantante}
\begin{algorithmic}[1]
\Require Number of agents $\boldsymbol{\Nagents}$;
local optimizers $\{\boldsymbol{\optimizer{\agentidx}}\}_{\agentidx=1}^{\Nagents}$;
scorable task $\boldsymbol{\task}$, with queries $\boldsymbol{\queryset}  = \{ \boldsymbol{\queryidx}_i\}_{i=1}^n$;
number of joint configurations $\boldsymbol{\Nconfigs}$;
group size $\boldsymbol{\groupsize}$;
attributer $\boldsymbol{\attributer}$;
number of iterations $\boldsymbol{\Niters}$;
\Ensure Updated optimizers $\{\boldsymbol{\optimizer{\agentidx}}\}_{\agentidx=1}^{\Nagents}$

\State $t \gets 0$
\While{$t < \Niters$}
  \For{$(\agentidx, \configidx) \in \{1,\dots,\Nagents\} \times \{1,\dots,\Nconfigs\}$}
    \State $\param{\agentidx}{\configidx} \gets \optimizer{\agentidx}.\Suggest()$
  \EndFor
  \For{$\configidx = 1$ \textbf{to} $\Nconfigs$}
    \State $\jointconfig{\configidx} \gets (\param{1}{\configidx}, \dots, \param{\Nagents}{\configidx})$ \Comment{\textcolor{olive}{Construct candidate joint configurations}}
  \EndFor

  \For{$(\configidx, \queryidx) \in \{1,\dots,\Nconfigs\} \times \queryset$}
    \State $(\trajectory{\queryidx}{\configidx},\, \score{\queryidx}{\configidx}) \gets \task.\Evaluate(\queryidx, \jointconfig{\configidx})$ \Comment{\textcolor{olive}{Evaluate joint configurations}}
  \EndFor

  \For{each $\queryidx \in \queryset$}
    \State Randomly partition $\{1,\dots,\Nconfigs\}$ into groups of size $\groupsize$
    \For{each group $\group{}$ in the partition} \Comment{\textcolor{olive}{Perform contrastive attribution}}
      \State $\{\credit{\queryidx}{\agentidx}{\configidx}\}_{\agentidx,\,\configidx} \gets \attributer.\Attribute\!\left(\queryidx,\, \{(\trajectory{\queryidx}{\configidx}, \score{\queryidx}{\configidx})\}_{\configidx \in \group{}}\right)$
    \EndFor
  \EndFor

  \For{$(\agentidx, \configidx) \in \{1,\dots,\Nagents\} \times \{1,\dots,\Nconfigs\}$}
    \State $\aggcredit{\agentidx}{\configidx} \gets \dfrac{1}{|\queryset|} \displaystyle\sum_{\queryidx \in \queryset} \credit{\queryidx}{\agentidx}{\configidx}$ \Comment{\textcolor{olive}{Aggregate credits per agent across queries}}
  \EndFor
  \For{$\agentidx = 1$ \textbf{to} $\Nagents$}
    \State $\optimizer{\agentidx}.\Update\!\left(\{(\param{\agentidx}{\configidx},\, \aggcredit{\agentidx}{\configidx})\}_{\configidx=1}^{\Nconfigs}\right)$
  \EndFor
  \State $t \gets t + 1$

\EndWhile
\State \Return $\{\optimizer{\agentidx}.\Suggest()\}_{\agentidx=1}^{\Nagents}$
\end{algorithmic}
\end{algorithm}

\section{Prompts} \label{app:prompts}
All prompts utilized in this work, including the full prompt sets for every optimizer, seed, and ablation study, as well as the alternative attribution prompt utilized in the additional experiments, are available in the supplementary material of this paper at \githuburl{}.

\subsection{Initial Prompt Creation} \label{app:pc}
We use the following Prompt~\ref{prompt:generation} to generate initial prompts from the agent's task descriptions. The placeholders for task description, input, and output variables are replaced with the respective agent-specific information before inference. We regenerate only outputs that violate the required interface, e.g., missing input variables, missing output tags, or malformed structured-output instructions.

\promptboxfile[prompt:generation]{Prompt Generation Prompt}{prompts/prompt_generation.txt}

\subsection{Prompt of the Attribution Model} \label{app:attributer_prompt}
Prompt~\ref{prompt:attributer} shows the full prompt used for the attribution model. The prompt instructs the attributer to compare multiple executions of the same query, each produced under a different agent parameterization, and to assign a credit score in $[-1, +1]$ to each agent per execution based on behavioral differences across executions, their effect on downstream reasoning, and the resulting system scores. To guide structured and consistent output, the prompt includes explicit formatting rules and a few-shot example covering three archetypal attribution scenarios: a fully correct execution, a failure originating in an early agent, and a localized formatting error. The placeholders for the number of parameterizations, the agent's name, and the rollouts are replaced with the respective runtime values before inference.
\promptboxfile[prompt:attributer]{Prompt of the Attribution Model}{prompts/attribution_prompt.txt}

\subsection{Initial Prompts} \label{app:init_prompts}
We report the initial prompt sets from seed 42 for each benchmark. All initial prompts are available in the paper's supplementary material at \githuburl{}.

Prompts~\ref{prompt:init_mbpp_planner}--\ref{prompt:init_mbpp_verifier} show the initial prompt set for MBPP corresponding to seed 42, which achieved \SI{7.0}{\%} on the respective evaluation set.

\promptboxfile[prompt:init_mbpp_planner]{Initial Prompts on MBPP: Planner Prompt}{prompts/init_mbpp_42/planner.txt}
\promptboxfile[prompt:init_mbpp_executor]{Initial Prompts on MBPP: Executor Prompt}{prompts/init_mbpp_42/coder.txt}
\promptboxfile[prompt:init_mbpp_verifier]{Initial Prompts on MBPP: Validator Prompt}{prompts/init_mbpp_42/verifier.txt}

Prompts~\ref{prompt:init_gsm8k_exec1}--\ref{prompt:init_gsm8k_cons} show the initial prompt set for GSM8K corresponding to seed 42, which achieved \SI{69.4}{\%} on the respective evaluation set.

\promptboxfile[prompt:init_gsm8k_exec1]{Initial Prompts on GSM8K: Executor 1 Prompt}{prompts/init_gsm8k_42/executor_1.txt}
\promptboxfile[prompt:init_gsm8k_exec2]{Initial Prompts on GSM8K: Executor 2 Prompt}{prompts/init_gsm8k_42/executor_2.txt}
\promptboxfile[prompt:init_gsm8k_exec3]{Initial Prompts on GSM8K: Executor 3 Prompt}{prompts/init_gsm8k_42/executor_3.txt}
\promptboxfile[prompt:init_gsm8k_cons]{Initial Prompts on GSM8K: Consensus Agent Prompt}{prompts/init_gsm8k_42/consensus.txt}

Prompts~\ref{prompt:init_hotpotqa_retriever}--\ref{prompt:init_hotpotqa_synthesizer} show the initial prompt set for HotpotQA corresponding to seed 42, which achieved \SI{7.2}{\%} on the respective evaluation set.

\promptboxfile[prompt:init_hotpotqa_retriever]{Initial Prompts on HotpotQA: Retriever Prompt}{prompts/init_hqa_42/retriever.txt}
\promptboxfile[prompt:init_hotpotqa_reader]{Initial Prompts on HotpotQA: Reader Prompt}{prompts/init_hqa_42/reader.txt}
\promptboxfile[prompt:init_hotpotqa_hallucination]{Initial Prompts on HotpotQA: Hallucination Detector Prompt}{prompts/init_hqa_42/hallucination.txt}
\promptboxfile[prompt:init_hotpotqa_synthesizer]{Initial Prompts on HotpotQA: Synthesizer Prompt}{prompts/init_hqa_42/synthesizer.txt}

\subsection{Best Prompts per Task} \label{app:best_prompts}
We report the highest-scoring prompt sets identified by \alg{} for each benchmark.

\subsubsection{MBPP} \label{app:best_prompts_mbpp}
Prompts~\ref{prompt:mbpp_planner}--\ref{prompt:mbpp_verifier} show the prompt set for MBPP corresponding to seed 47, which achieved \SI{47.4}{\%} on the respective evaluation set.

\promptboxfile[prompt:mbpp_planner]{\alg Prompts on MBPP: Planner Prompt}{prompts/cantante_mbpp_47/planner_prompt.txt}
\promptboxfile[prompt:mbpp_executor]{\alg Prompts on MBPP: Executor Prompt}{prompts/cantante_mbpp_47/coder_prompt.txt}
\promptboxfile[prompt:mbpp_verifier]{\alg Prompts on MBPP: Validator Prompt}{prompts/cantante_mbpp_47/validator_prompt.txt}

\subsubsection{GSM8K} \label{app:best_prompts_gsm8k}
Prompts~\ref{prompt:gsm8k_executor1}--\ref{prompt:gsm8k_consensus} show the prompt set for GSM8K corresponding to seed 7, which achieved \SI{85.6}{\%} on the respective evaluation set.
\promptboxfile[prompt:gsm8k_executor1]{\alg Prompts on GSM8K: Executor 1 Prompt}{prompts/cantante_gsm8k_7/executor_1_prompt.txt}
\promptboxfile[prompt:gsm8k_executor2]{\alg Prompts on GSM8K: Executor 2 Prompt}{prompts/cantante_gsm8k_7/executor_2_prompt.txt}
\promptboxfile[prompt:gsm8k_executor3]{\alg Prompts on GSM8K: Executor 3 Prompt}{prompts/cantante_gsm8k_7/executor_3_prompt.txt}
\promptboxfile[prompt:gsm8k_consensus]{\alg Prompts on GSM8K: Consensus Prompt}{prompts/cantante_gsm8k_7/consensus_prompt.txt}

\subsubsection{HotpotQA} \label{app:best_prompts_hotpotqa}
Prompts~\ref{prompt:hotpotqa_retriever}--\ref{prompt:hotpotqa_synthesizer} show the prompt set for HotpotQA corresponding to seed 7, which achieved \SI{17.4}{\%} on the respective evaluation set.
\promptboxfile[prompt:hotpotqa_retriever]{\alg Prompts on HotpotQA: Retriever Prompt}{prompts/cantante_hqa_7/retriever_prompt.txt}
\promptboxfile[prompt:hotpotqa_reader]{\alg Prompts on HotpotQA: Reader Prompt}{prompts/cantante_hqa_7/reader_prompt.txt}
\promptboxfile[prompt:hotpotqa_det]{\alg Prompts on HotpotQA: Hallucination Detector Prompt}{prompts/cantante_hqa_7/hallucination_prompt.txt}
\promptboxfile[prompt:hotpotqa_synthesizer]{\alg Prompts on HotpotQA: Synthesizer Prompt}{prompts/cantante_hqa_7/synthesizer_prompt.txt}

\end{document}